\documentclass[lettersize,journal]{IEEEtran}
\usepackage{amsmath,amsfonts}
\usepackage{algorithmic}
\usepackage{algorithm}
\usepackage{array}
\usepackage[caption=false,font=normalsize,labelfont=sf,textfont=sf]{subfig}
\usepackage{textcomp}
\usepackage{stfloats}
\usepackage{url}
\usepackage{verbatim}
\usepackage{graphicx}
\usepackage{cite}
\usepackage[utf8]{inputenc}
\usepackage[T1]{fontenc}
\usepackage{layouts}
\usepackage{setspace}
\usepackage{longtable}
\usepackage[version=4]{mhchem}       
\usepackage{multirow}
\usepackage{array}
\usepackage{hyperref}
\usepackage[version=4]{mhchem}
\usepackage{xcolor}
\usepackage{xr}
\usepackage{cleveref}
\usepackage{longtable}
\usepackage{multicol}

\usepackage{listings}
\definecolor{light-gray}{rgb}{0.9,0.9,0.9}

\newcolumntype{P}[1]{>{\centering\arraybackslash}p{#1}}

\begin{document}

\title{Predicting Soil Macronutrient Levels: A Machine Learning Approach Models Trained on pH, Conductivity, and Average Power of Acid-Base Solutions}

\author{Mridul Kumar$^{1*}$, Deepali Jain$^2$, Zeeshan Saifi$^2$, Krishnananda Soami Daya$^{2*}$ \\
$^*$Corresponding Authors \\
\textit{$^1$Department of Materials Engineering, Ben-Gurion University of the Negev, Be'er Sheva, Israel}\\
\textit{$^2$Department of Physics and Computer Science, Dayalbagh Educational Institute, Agra, India} \\
Email : *\href{mailto:mridul@post.bgu.ac.il}{mridul@post.bgu.ac.il}, *\href{mailto:ksdaya@dei.ac.in}{ksdaya@dei.ac.in}
}


\onecolumn

\maketitle
\begin{abstract}
Soil macronutrients, particularly potassium ions (K$^+$), are indispensable for plant health, underpinning various physiological and biological processes, and facilitating the management of both biotic and abiotic stresses. Deficient macronutrient content results in stunted growth, delayed maturation, and increased vulnerability to environmental stressors, thereby accentuating the imperative for precise soil nutrient monitoring. Traditional techniques such as chemical assays, atomic absorption spectroscopy, inductively coupled plasma optical emission spectroscopy, and electrochemical methods, albeit advanced, are prohibitively expensive and time-intensive, thus unsuitable for real-time macronutrient assessment. In this study, we propose an innovative soil testing protocol utilizing a dataset derived from synthetic solutions to model soil behaviour. The dataset encompasses physical properties including conductivity and pH, with a concentration on three key macronutrients: nitrogen (N), phosphorus (P), and potassium (K). Four machine learning algorithms were applied to the dataset, with random forest regressors and neural networks being selected for the prediction of soil nutrient concentrations. Comparative analysis with laboratory soil testing results revealed prediction errors of 23.6\% for phosphorus and 16\% for potassium using the random forest model, and 26.3\% for phosphorus and 21.8\% for potassium using the neural network model. This methodology illustrates a cost-effective and efficacious strategy for real-time soil nutrient monitoring, offering substantial advancements over conventional techniques and enhancing the capability to sustain optimal nutrient levels conducive to robust crop growth. 
\end{abstract}


\section{Introduction}
Macronutrients are indispensable elements required by plants in substantial quantities, playing a pivotal role in diverse physiological and biochemical processes, including photosynthesis, respiration, and the synthesis of proteins and amino acids \cite{kimSoilMacronutrientSensing2009}. The primary macronutrients encompass nitrogen (N), phosphorus (P), potassium (K), calcium (Ca), magnesium (Mg), and sulphur (S) \cite{hawkesford2023functions}. Each macronutrient fulfils distinct functions; for instance, nitrogen is a crucial constituent of amino acids and is imperative for the formation of plant tissues. It is also an integral part of nucleic acids and chlorophyll \cite{lewis1986plants}. Phosphorus is vital in critical processes such as energy transfer via adenosine triphosphate (ATP), photosynthesis, and respiration. It is also a component of phospholipids and nucleic acids, which are essential for cell membrane integrity \cite{raghothama2005phosphorus}. Phosphorus is particularly significant for root development, flowering, fruiting, and seed formation. Adequate phosphorus availability enhances the plant's ability to capture and utilize solar energy efficiently. Potassium is instrumental in enzyme activation, osmoregulation, and the synthesis of proteins and starches. It modulates the opening and closing of stomata, thereby regulating water use efficiency and transpiration \cite{amtmann2012potassium}. Potassium augments disease resistance, fortifies drought tolerance, and contributes to the overall resilience of plants \cite{hasanuzzamanPotassiumVitalRegulator2018}. Additionally, it is essential for the transport of sugars and other nutrients within the plant. Calcium serves as a structural component of cell walls and membranes, facilitating cell division, stability, and integrity. Magnesium, being the central atom in the chlorophyll molecule, is indispensable for photosynthesis \cite{white2003calcium}.

Deficiencies in macro-nutrients can lead to inhibited plant growth and substantially alter the plant's response to external stressors \cite{wangCriticalRolePotassium2013}. For instance, nitrogen deficiency is distinguished by stunted growth and chlorotic, pale yellow leaves. The overall vigour of the plant diminishes, and there is a pronounced reduction in yield owing to impaired photosynthesis and protein synthesis \cite{mu2021physiological}. Phosphorus deficiency results in dark green or purplish foliage and delayed plant maturity. Such plants also demonstrate reduced energy transfer, which adversely affects overall growth and productivity \cite{aziz2014phosphorus}. Potassium deficiency presents as chlorosis along the leaf margins, followed by necrosis of leaf tissue. Consequently, plants exhibit increased susceptibility to drought and disease, and the structural integrity of the stem is impaired \cite{machinandiarenaPotassiumPhosphitePrimes2012,wangCriticalRolePotassium2013}. Calcium deficiency results in suboptimal root development and compromised cell wall integrity, leading to distorted growth. Magnesium deficiency induces interveinal chlorosis in older leaves, where the tissue between veins becomes yellow while the veins remain green, thereby reducing photosynthetic efficacy and overall growth. This cumulative decline in plant health engenders a reduction in crop quality and quantity by approximately 30\% \cite{oerkeCropLossesPests2006,kumarDecodingPhysiologicalResponse2023}. Given projections that the global population will reach 10 billion by 2050 \cite{bureauTotalMidyearPopulation2008}, it is crucial to address and mitigate these detriments in crop quality and yield.
Consequently, it is imperative to meticulously monitor macronutrient levels in the soil, as their availability is modulated by variables such as soil composition, climatic conditions, and historical land use. Current methodologies for soil nutrient detection encompass colorimetric techniques, which involve chemical reagents that interact with specific nutrients to induce a quantifiable colour change, thereby ascertaining nutrient concentrations \cite{jones2001laboratory}. Moreover, advanced spectroscopic methods, including atomic absorption spectroscopy (AAS), X-ray fluorescence spectrometry (XFS), and inductively coupled plasma optical emission spectroscopy (ICP-OES), are utilized for precise nutrient quantification in soil \cite{khairaInductivelyCoupledPlasma2011,nemesXrayFluorescenceSpectrometry2015}. Additionally, electrochemical approaches such as ion-selective electrodes are employed to measure specific ion concentrations in soil solutions \cite{kim2007simultaneous}. Despite their sophistication, these techniques are frequently cost-prohibitive and time-intensive. Furthermore, by the time agronomists receive soil testing reports, macronutrient levels may have already shifted, thereby rendering these methods inadequate for real-time monitoring. These constraints highlight the exigency for a more economical, rapid, and efficient soil nutrient analysis technique, thereby incentivizing the development of innovative methods for nutrient detection.

The methodology delineated in this study addresses the physical properties of soil, specifically conductivity and pH, utilizing machine learning regression techniques to both qualitatively and quantitatively evaluate the concentrations of macronutrients. Given that machine learning models necessitate a dataset to discern meaningful patterns, it was essential to develop a dataset encompassing soil physical properties alongside the corresponding macronutrient concentrations. Nonetheless, soil testing is a protracted procedure. Consequently, we synthesized acid-base solutions with predetermined concentrations of the respective acids and bases of the macronutrients to replicate the soil characteristics. For simplicity, we focused on three macronutrients: nitrogen, phosphorus, and potassium. The acids and bases used for these macronutrients included \ce{HNO3}, \ce{H3PO4}, and \ce{KOH}.

Four distinct machine learning regression methodologies—specifically, linear regression, random forest, k-nearest neighbours (KNN), and neural networks—were deployed to forecast soil nutrient concentrations. The investigation targeted the estimation of \ce{HNO3}, \ce{H3PO4}, and \ce{KOH} levels. Among these methodologies, neural networks and random forest regressors exhibited superior predictive accuracy for nutrient concentrations. Consequently, these two models were selected for subsequent analysis using empirical soil samples. To substantiate the precision of the trained models, 10 soil samples were procured from two separate locations and subjected to standard soil testing protocols to ascertain the actual concentrations of \ce{HNO3}, \ce{H3PO4}, and \ce{KOH}. The laboratory findings were subsequently juxtaposed with the predictions derived from the neural network and random forest models. This comparative analysis facilitated the evaluation of the models' reliability and accuracy in predicting soil nutrient levels in real-world conditions.

\section{Materials and Methods}

Soil is a complex and dynamic mixture of organic matter, minerals, water, air, and living organisms. It forms a critical interface between the land and the atmosphere and plays a crucial role in supporting plant growth and sustaining terrestrial ecosystems. Soil fertility, or the ability of a soil to support plant growth, is dependent on a range of factors, including soil nutrients, pH, and organic matter content \cite{schlesingerBiogeochemistryAnalysisGlobal2013}. The major soil nutrients which are required for plant growth are Nitrogen (N), Phosphorus (P), and Potassium (K), These nutrients are generally referred to as macro-nutrients as they are required in high quantities by the plants \cite{marschnerMineralNutritionHigher1995}. These nutrients are generally found as chemical compounds in the soil, however, in the presence of water get converted to their ionic forms (see Table \ref{tab:soil_nutrients} and see equations \ref{eq:hno3}, \ref{eq:h3po4}, and \ref{eq:koh}). If X$^{n+}$ is a cation with $n$ positive charge and Y$^{m-}$ is the anion in the chemical compounds, then in the presence of water the decomposition to ionic forms can be given as,

\begin{center}
\begin{equation}
	\ce{X(NO3)_n ->[Aqueous] X^{n+} + nNO^-_3}
	\label{eq:hno3}
\end{equation}
\begin{equation}
	\ce{X3(PO4)_n ->[Aqueous] 3X^{n+} + nPO^{3-}_4}
	\label{eq:h3po4}
\end{equation}
\begin{equation}
	\ce{K_mY ->[Aqueous] mK+ + Y^{m-}}
	\label{eq:koh}
\end{equation}
\end{center}

It is apparent from the above equations that the chemical compounds of three primary macro-nutrients (N, P, and K) break down to form the ions which are then absorbed by the plants. For a proper growth and development of the plant, appropriate amounts of these nutrients is required. For this purpose, farmers generally get their soil tested \cite{khairaInductivelyCoupledPlasma2011,olsen1954estimation}. However, these methods are expensive and time-consuming, and they do not provide real-time information on soil nutrient concentrations. As a result, there is a risk of inaccurate soil assessments, which can lead to inappropriate fertilizer usage. Therefore, there is a need of a soil testing technique which can tell the real-time information of the concentration of soil nutrients.

Rise of machine learning methods in different fields such as for plant disease detection, classification, and predictive modelling led us to believe that this technique can also be applied for the prediction of soil nutrients on the basis of soil physical parameters \cite{kumarDecodingPhysiologicalResponse2023, garcia-pedrajasEmpiricalStudyBinary2011, kumarNeuralNetworksClassroom2004}. However, collecting soil and creating a database consisting of its physical properties and known concentration of nutrients is arduous and time-consuming as we would have to go through the soil testing methods. Therefore, we have taken the bottom-up approach by creating the soil phantoms using the known concentrations of the constituting ions (in their acid and base form). In this paper, we only focus on the three macro-nutrients (N, P, and K) in their ionic forms as acids and base i.e. \ce{HNO3}, \ce{H3PO4}, and \ce{KOH}. Three physical properties namely, solution pH, conductivity, and average power were calculated for the soil phantoms and were put in a dataset with the known concentrations of the constituting acids and base.

\subsection{Preparation of the Solutions}
Respective acids and base of the macro-nutrients found in the soil were collected from the Department of Chemistry. The concentrations of these chemicals were noted from the labels. For making the dataset, we have to take the concentration of the acids and bases such that it is equivalent to the concentrations of the macro-nutrients found in the soil so that the machine learning models can predict that level of the concentration. However, if the concentration in the dataset is different from the concentration of the macro-nutrients found in the soil, then the machine learning models will not be generalized; hence, the output will not be appropriate on soil data. The concentration of the macro-nutrients found in the soil can be seen in Table \ref{tab:soil_nutrients}.

The concentration of \ce{H3PO4} was 1.7 g/mL and the concentration of \ce{HNO3} was 1.42 g/mL These acids were diluted using the equation \ref{eq:dilution}, where $ M_1 $ and $ V_1 $ are the molarity and volume respectively of the initial solution and $ M_2 $ and $ V_2 $ are the molarity and volume respectively of the final solution,
\begin{equation}
	M_1V_1 = M_2V_2
	\label{eq:dilution}
\end{equation}
KOH is available as the palettes and a solution of known molar concentration ($ M $) of it was prepared using equation \ref{eq:koh_sol}.
\begin{equation}
	M = \frac{m}{56.1 \times V} \times 1000
	\label{eq:koh_sol}
\end{equation}
Where, $ m $ is the mass of the $ KOH $ palettes in grams and V is the volume of the final solution in mL and 56.1 gm/mol is the molar mass of the $ KOH $. It is to be noted that the prepared solution of $ KOH $ can further be diluted using the equation \ref{eq:dilution}.

\subsubsection{Nitric Acid (\boldmath{$ HNO_3 $})} 
When mixing acids with water, it's important to add the acids to the water, not the other way around. This precaution is necessary because the mixing process is exothermic and may cause the acids to spill out. Additionally, it's advisable to add the acids in small quantities whenever possible. In our case, the concentration of \ce{HNO3} was 1.42 g/mL. We measured 210 mL of water into a 250 mL beaker. Using a micropipette, we added 739 µL of \ce{HNO3} to the water and carefully stirred the solution. This resulted in a final solution with a concentration of 0.08 M, equivalent to 5040 ppm.

\subsubsection{Ortho-phosphoric Acid (\boldmath{$ H_3PO_4 $})}
The concentration of the original solution of $ H_3PO_4 $ in our case was 1.7 g/mL which was noted down from the bottle label. In another 250 mL beaker, 210 mL water was poured for performing the dilution. Using a different micropipette, 62 µL concentration of the $ H_3PO_4 $ was taken and dropped into the beaker. The final solution was carefully stirred, and its concentration was 0.005 M, which is 490 ppm.

\subsubsection{Potassium Hydroxide (\boldmath{$ KOH $})}
We wanted to make a potassium hydroxide solution of 0.535 M concentration, which is close to 30000 ppm. For this purpose, 6.3 gm of \ce{KOH} palettes were weighed to be mixed with water for making a final solution of 210 mL volume, which is 30000 ppm.

\subsection{Mixing of the Prepared Solutions}
The prepared solutions were mixed in various proportions to create a final solution for which we measured the pH, electrical conductivity, and V-I characteristic. This was done by placing the solution in a beaker as shown in Figure \ref{fig:setup_exp5}C. We kept the volume of the final solution fixed at 40 mL to measure these physical parameters at different concentrations of the acids and bases in different proportions.

We began by mixing 40 mL of 0.535 M \ce{KOH}, gradually increasing the volume of the other two acids in 2 mL increments. Following this method, we were able to prepare 231 solutions, each with a final volume of 40 mL, in which each acid and base was present in different proportions.

\subsection{Measurement of Physical Parameters}
Solution pH, conductivity, and area under V-I characteristics curve (average power) were measured. Solution pH shows which kind of ions (cations or anions) are higher in the solution, conductivity tells us if the solution is composed of lower or higher conduction ions. Finally, V-I characteristics curve helps us in understanding the conduction of current at different voltages.

\subsubsection{pH of the Solution}
pH of the solution is the negative logarithm of the hydrogen ion concentration in the solution. When the solution has higher amount of acids (\ce{NO3-} and \ce{PO_4^{3-}} ions in our case) the pH of the solution decreases and when there are basic compounds in the solution (\ce{KOH} in our case) its pH increases.
\begin{equation}
\text{pH} = -\log_{10}H^+
\end{equation}
Therefore, a machine learning model would be able to learn if the solution is made up of acidic or basic ions.

\subsubsection{V-I Characteristics Curve}
V-I characteristics curve has been calculated using the experimental setup shown in Figure \ref{fig:setup_exp5}D. In this setup, data acquisition software is created with Python programming language. This software controls the programmable DC power supply and digital multimeter. It starts by first setting voltage 0 V at the output of programmable DC power supply then it increases the output voltage in intervals of 50 mV and measures the current through the circuit using the digital multimeter and stores all this information in a comma separated value (CSV) file with the known concentrations of acids and base.
We have calculated two physical parameters \textit{i.e.} electrical conductivity and average power through the solution using the following formulae.

\begin{equation}
	P_{av} = \int_{0}^{5}IdV
	\label{eq:power}
\end{equation}

\begin{equation}
	\sigma = \frac{l}{RA}
	\label{eq:conductivity}
\end{equation}

In the above equation, $ R $ is the electrical resistance of the solution and can be given as $ R = \frac{dV}{dI}$, $ l $ is the separation between the electrodes ($ l = 4.5 cm $), and $ A $ is the area of the electrode immersed in the solution ($ A = 1.26 cm^2$).

Since, we are dealing with a discrete data here, Simpson's 3/8$^{\text{th}}$ rule was applied to the data collected for calculating the integral in equation \ref{eq:power}, and equation \ref{eq:conductivity} was converted to the following,

\begin{equation}
	\sigma = \frac{dI}{dV} \frac{l}{A}
\end{equation}
The above value of the conductivity will be calculated at each point of the sampling at the gap of $ dV = 0.05V $, therefore, this conductivity will have to be averaged for all the sampled values and the final equation will become, 
\begin{equation}
	\begin{split}
	\sigma (S/m) & = \frac{1}{100}\sum_2^{101} \frac{0.045}{0.05 \times 0.000126} (I_n - I_{n - 1}) \\
	& = \sum_2^{101} 71.42 (I_n - I_{n - 1})
	\end{split}
\end{equation}
The electrical force felt by the ions with charge $ q $ and mass $ m $ can be given as,
\begin{equation}
	\vec{F} = q\vec{E}
\end{equation}
From the above equation the acceleration felt by the charge carrier can also be calculated,
\begin{equation}
	\vec{a} = \frac{q\vec{E}}{m}
\end{equation}
This acceleration would cause the charge carriers or the ions in the solution to drift from one electrode to the other causing the current to flow. The rate of change of position can also be given as,
\begin{equation}
	\frac{d^2\vec{x}}{dt^2} = \frac{q\vec{E}}{m}
	\label{eq:acc_charge}
\end{equation}
The magnitude of the electric field can be given as, $ E = \frac{V}{l} $. Let us assume that the particle starts from one electrode, so at $ t = 0 $, $ x = 0 $. So, the final equation of change in the position of the particle can be given as,
\begin{equation}
	x = \frac{qVt^2}{ml}
\end{equation}
Since, we have kept $ V $ and $ l $ constant for all the solutions, the rate of change of position with time will only dependent upon $ q $ and $ m $. This would help machine learning models in identifying which ion is in large concentration in the solution.

\subsection{Creation of the Dataset and Training of the Models}
A separate Python program was written using the formulae discussed in previous subsections (see code \ref{code:imp1}). This program created the acid-base dataset in which (pH, conductivity ($ \sigma $), and average power ($P_{av} $)) were the input features and the concentration of \ce{HNO3}, \ce{H3PO4}, and \ce{KOH} were the output features.

These input features are of different order e.g. pH lies between 0-15, but, conductivity in our case lies between 0.02 - 5.34 S/m. Therefore, there is a need to scale the data so that all the input features have mean ($ \mu $) of 0 and standard deviation ($ SD $) of 1. This would avoid the model being too sensitive for change in one feature. To achieve this outcome \textit{StandardScalar()} class from \textit{scikit-learn} library was used to scale the input features before training the machine learning models.

Four different machine learning regression models were trained and tested on the dataset. These were linear regressor, k-nearest neighbour, random forest, and neural network. These models were trained and tested on the dataset and their accuracy was measured based on the mean absolute error (MAE). These results were compared to decide two best performing algorithms for real world prediction of soil macronutrients (N, P, and K). 

To mitigate overfitting and enhance model generalization, we employed k-fold cross-validation technique on the dataset. With a k-value of 5, the dataset was partitioned into five equal segments. The model underwent iterative training on four segments, while performance was evaluated on the remaining segment for generalization. This process was repeated, ensuring each segment served as the validation set once. 

\begin{table}
\setlength{\tabcolsep}{2pt} 
\renewcommand{\arraystretch}{1.5} 
\caption{Typical range of concentration of nutrients in soil \cite{eash2015soil,jones2012plant}.}
\label{tab:soil_nutrients}
\centering
	\begin{tabular}{|c|c|c|}
		
		\hline
		\textbf{Nutrient} & \textbf{Typical Range} & \textbf{Ions Available in Soil} \\ \hline
		Nitrogen (N) & 500 - 5,000 ppm & \ce{NO^-3}, \ce{NO^-2}, \ce{NH^+4}  \\ \cline{1-3}
		Phosphorus (P) & 50 - 500 ppm & \ce{PO^{3-}_4}, \ce{H2PO4^-}, \ce{HPO4^2-}  \\ \cline{1-3}
		Potassium (K) & 1,000 - 30,000 ppm & \ce{K+}  \\ \cline{1-3}
		Calcium (Ca) & 1,000 - 50,000 ppm  & \ce{Ca^2+}  \\ \cline{1-3}
		Magnesium (Mg) & 500 - 10,000 ppm & \ce{Mg^2+}  \\	\cline{1-3}
		Sulphur & 1,000 - 5,000 ppm & \ce{SO4^2-}  \\ \cline{1-3}
	\end{tabular}
\end{table}
\subsection{Application of Machine Learning Models on Soil}
\begin{figure*}
	\centering
	\includegraphics[width=14cm]{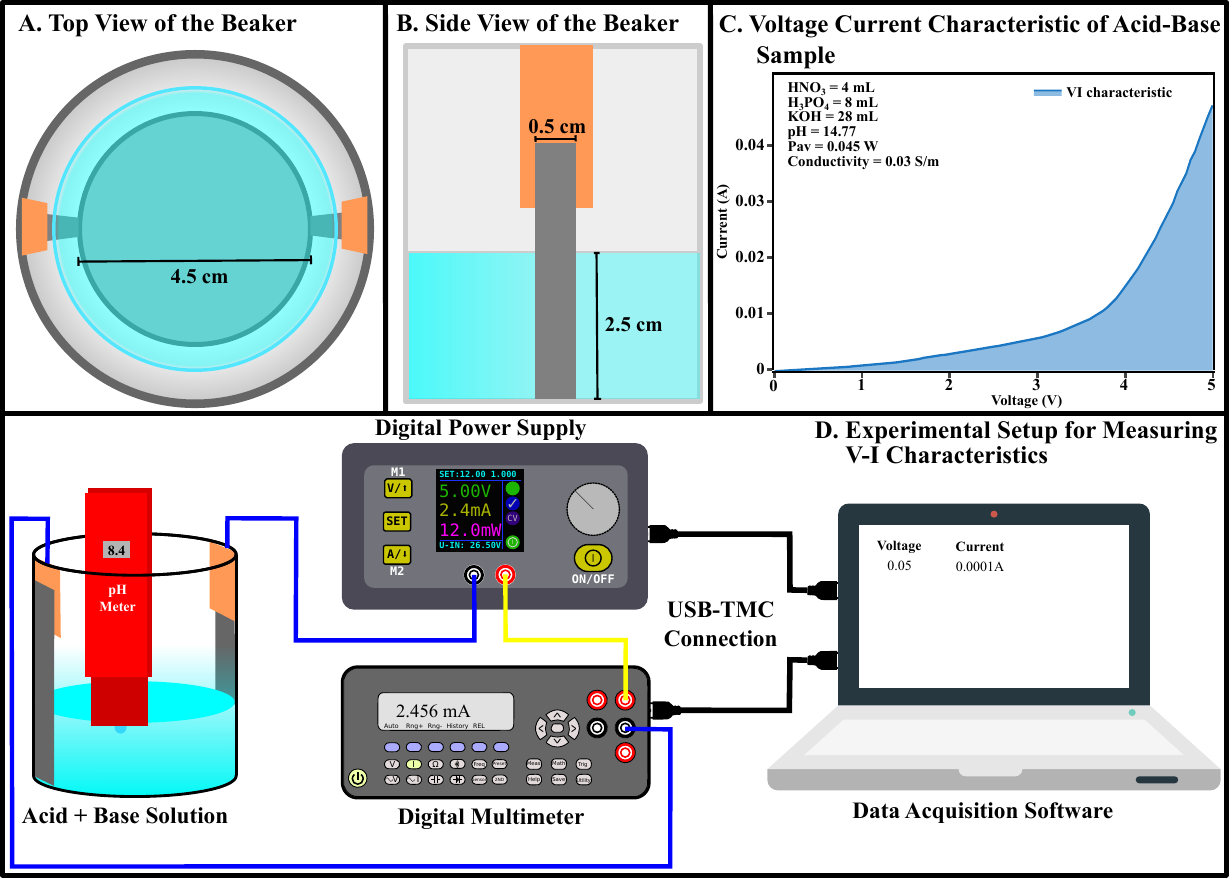}
	\caption{This figure shows the preparation of the experimental setup. A. Top view of the beaker in which characteristics of final solution is measured. B. Side view of the beaker with the final volume of the solution. C. V-I characteristics of the final solution, the shaded area shows the average power transferred through the solution. D. The experimental setup for measuring the pH and calculating the V-I characteristics of the acid-base solutions.}
	\label{fig:setup_exp5}
\end{figure*}

10 samples of soil were collected from two different places for testing the concentration of the selected macro-nutrients in it using the proposed method. An amount of 5 gm each was put in beakers and water was carefully poured to reach the final volume of 50 mL. These solutions were thoroughly mixed to make a colloidal solution. The final solution was filtered to reach a final volume of 40 mL. For testing the electrical properties of the sample, we poured it in the testing beaker of the experimental setup as shown in Figure \ref{fig:setup_exp5}. After that pH of the solution was measured. Conductivity of the solution was calculated using the equation \ref{eq:conductivity}. These tests were repeated for all 10 samples and a csv file was prepared to feed the data into machine learning models for prediction.

We have trained our models on the ionic concentration (mmol/L) of the respective macro-nutrients, therefore, the predicted concentration of the macro-nutrients is in mmol/L. However, the standard techniques use kg/ha as the standard unit for calculating the macronutrient concentration, our data needs to be converted to the standard values. For this purpose, the bulk density of the soil is required, the average soil density in Agra, Uttar Pradesh is 1.5 g/cm$ ^3 $ \cite{mustafa2011characterization}. If $ x $ is the molar concentration of the nutrients (in mmol) and $ M $ is the molar mass (in g/mol) then the total mass of the nutrient in 1 L of the solution can be given as,
\begin{equation}
	m = x \times M \quad \text{(in mg/L equivalent to ppm)}
\end{equation}
This mass ($ m $) would be available in 1 L of soil solution. For finding out the total mass in one hectare of soil, we first have to calculate the mass of it. Let the depth of the soil be $ d $ and area be $ A $ and bulk density be $ \rho $, then the mass of the soil can be given as,
\begin{equation}
	m_{soil} = \rho \times d \times A
\end{equation}
In our case, $ A (1 ha) = 10^4  m^2 $ and $ d = 15cm $ and as previously discussed, $ \rho = 1.5 g/cm^3 $. Therefore, the mass of the soil will be,
\begin{equation}
	m_{soil} = 1500 kg/m^3 \times 0.15 m \times 10^4 = 2.25 \times 10^6 kg/ha
\end{equation}
Therefore, the concentration of the nutrients in kg/ha can be given as,
\begin{equation}
	conc (kg/ha) = m \times 2.25 \quad kg/ha 
\end{equation}
Therefore, in our case the concentration of the soil can be converted to kg/ha by multiplying the concentration of the nutrients in the soil (in mg) by 2.25. It is to be noted here that the concentration of the soil nutrients is measured in the form of chemical compounds. For an example, potassium is measured in the form of \ce{K2O} \cite{pauline1946flame} and phosphorus is measured in the form of \ce{P2O5} \cite{olsen1954estimation}. However, our methods predict the concentration of the nutrients in the form of \ce{KOH}, \ce{H3PO4}, and \ce{HNO3}. Therefore, our predicted concentration of chemical compounds has to be converted to standard compounds. This can be done using the following chemical equations.
\begin{equation}
\ce{2KOH -> K2O + H2O}
\label{eq:koh_k2o}
\end{equation}
From equation \ref{eq:koh_k2o}, it is clear that 2 mols of \ce{KOH} make up 1 mol of \ce{K2O}. Therefore, the concentration of \ce{K2O} would be half of the \ce{KOH}. Similarly,
\begin{equation}
\ce{2H3PO4 -> P2O5 + 3H2O}
\label{eq:h3po4_p2o5}
\end{equation}
It is clear from equation \ref{eq:h3po4_p2o5} that 2 mols of \ce{H3PO4} make 1 mol of \ce{P2O5}. Therefore, the amount of \ce{P2O5} will also be half of \ce{H3PO4}.

We have neglected Nitrogen as \ce{HNO3} here because we did not get the measurements of it from the soil testing lab.

\section{Results}

\subsection{Empirical Examination of Feature Interdependencies within the Dataset}
\begin{figure}[t]
	\centering
	\includegraphics[width=7cm]{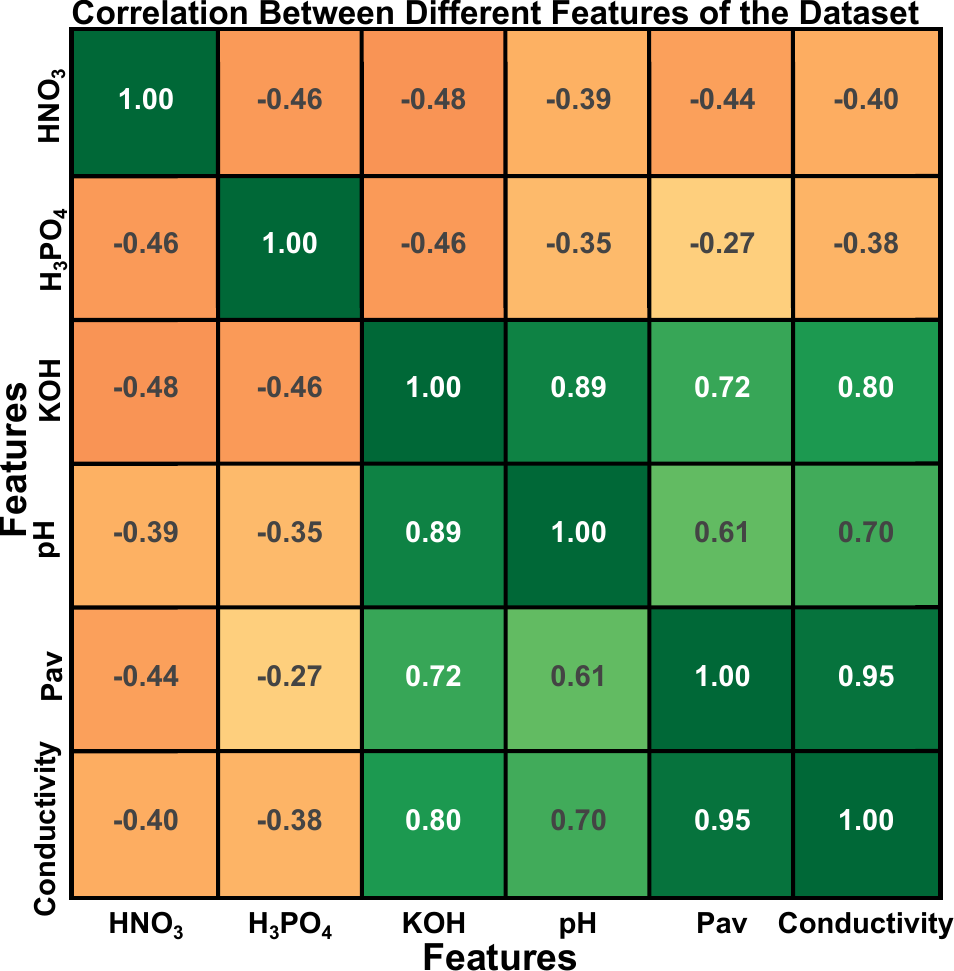}
	\caption{This figure illustrates the Spearman correlation coefficient (SCC) among the dataset's features, including acid-base concentration, pH, P$_{av}$, and conductivity. The elements along the diagonal (from the top-left to the bottom-right) represent the self-correlation of each feature and are consequently non-informative for the analysis..}
	\label{fig:feature_correlation}
\end{figure}

Prior to initiating the training module, it was crucial to scrutinize the interdependencies among the various features within the dataset. This was accomplished by computing the Spearman correlation coefficient between the different features (refer to Figure \ref{fig:feature_correlation}). It was observed that an increase in the concentration of a particular compound generally precipitates a decrease in the concentrations of the other two. Consequently, the concentrations of the acids (\ce{H3PO4} and \ce{HNO3}) and the base (\ce{KOH}) exhibit a negative correlation with each other. Furthermore, potassium ions (\ce{K+}) are well-documented for their proficient conductivity in aqueous solutions and their basic characteristic. As such, the concentration of potassium hydroxide (\ce{KOH}) is strongly positively correlated with the solution's conductivity, pH level, and mean power output. Conversely, nitric acid (\ce{HNO3}) and phosphoric acid (\ce{H3PO4}) dissociate in water to yield the negatively charged ions nitrate (\ce{NO3-}) and phosphate (\ce{PO4^3-}), respectively. While an increase in the concentration of these ions augments conductivity and average power, they are comparatively denser and less conductive than \ce{K+} ions (see Equation \ref{eq:acc_charge}). Thus, a relative diminution in conductivity and average power is apparent as the concentration of these negative ions amplifies. Additionally, these negative ions exhibit acidic properties, resulting in an inverse correlation between the concentrations of \ce{HNO3} and \ce{H3PO4} and the solution's pH level.
Conductivity and average power indeed manifest a discernible positive correlation. This relationship can be ascribed to the concomitant variations observed in these parameters. Moreover, the solution's pH is elevated by an augmentation in the concentration of \ce{KOH}, while it is diminished by a rise in the concentrations of \ce{HNO3} and \ce{H3PO4}. Consequently, both conductivity and average power exhibit a positive correlation with pH.

\subsection{Training of the Models}
Four distinct machine learning regression algorithms were selected for the analysis of the acid-base dataset, specifically linear regression, k-nearest neighbours (k-NN), random forest, and neural networks. To enhance generalizability and mitigate overfitting, a 5-fold cross-validation method was employed. Additionally, to further augment the model accuracy, principal component analysis (PCA) was implemented on both the training and testing datasets.

\subsubsection{Linear Regressor}
A linear regressor constitutes the most elementary model for forecasting outputs based on input features. This algorithm endeavours to align a linear trajectory within the feature space, aiming to minimize the prediction error, specifically the mean absolute error (MAE) in this context. Denoting the output as $ y $, the features as $ X_i $, and a constant as $ c $, the relationship can be described as follows \cite{montgomery2021introduction},
\begin{equation}
	y = \sum_{1}^{n}m_i X_i + c
\end{equation} 
\begin{figure*}[!h]
	\centering
	\includegraphics[width=14cm]{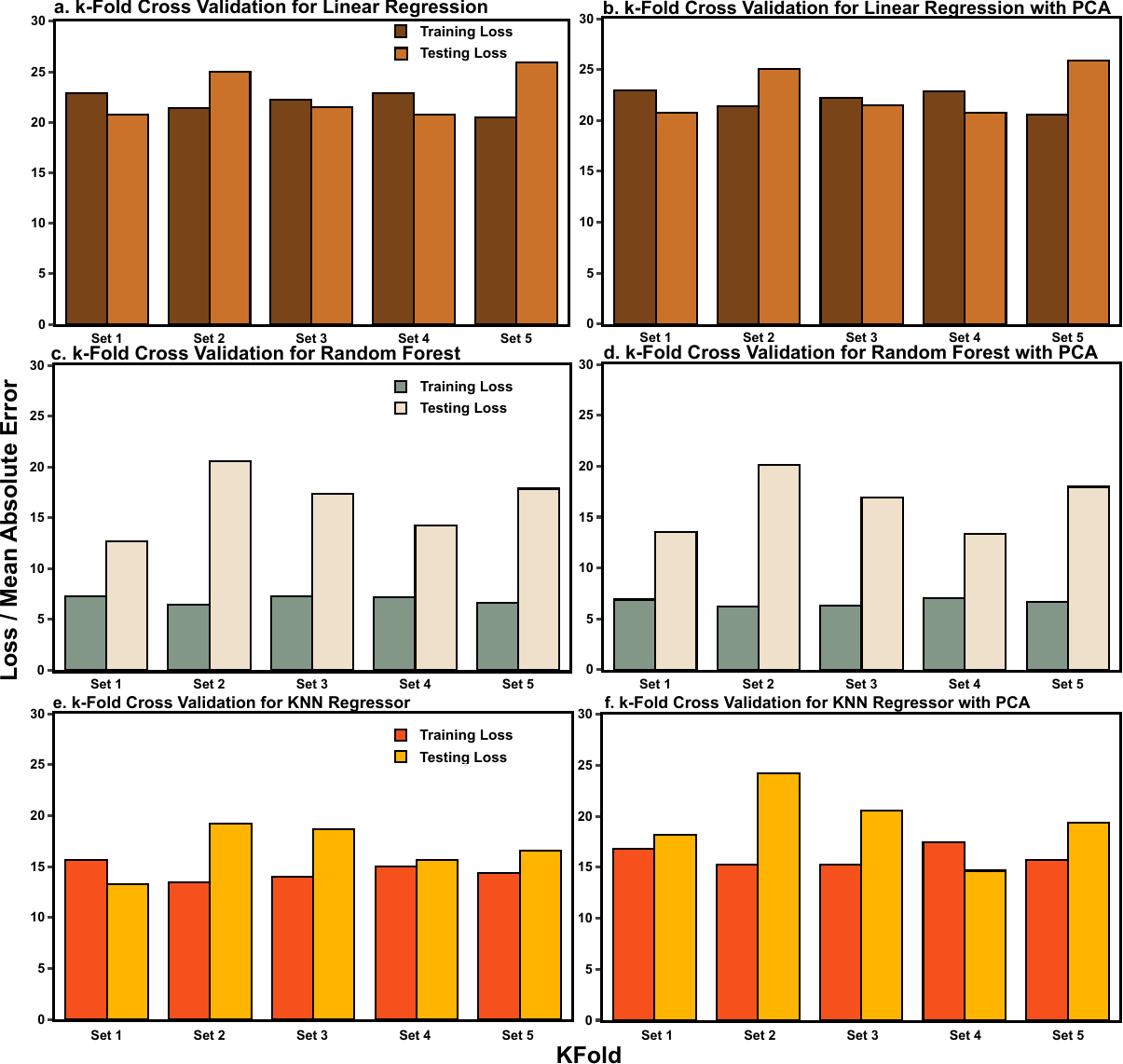}
	\caption{MAE loss on different kfold set for all the algorithms used. a. Linear model without PCA. b. Linear model with PCA. c. Random forest without PCA. d. Random forest with PCA. e. k-NN regressor without PCA. f. k-NN regressor with PCA.}
	\label{lrcomp}
\end{figure*}
\textit{LinearRegression} was used from \textit{sklearn.linear\_model} library. The hyperparameters used for training and testing are shown in the following code snippet,

\begin{lstlisting}{language = Python}
regressor = LinearRegression(fit_intercept=True,
			copy_X=True,
			n_jobs=None,
			positive=False)
\end{lstlisting}
Training MAE without application of PCA in this case was 22.9, 21.4, 22.2, 22.9, 20.5 and testing MAE was 20.8, 25, 21.5, 20.8, 25.9 for set 1-5 respectively. After the application of PCA, the training MAE was 22.9, 21.4, 22.2, 22.9, 20.5 and the testing MAE was 20.8, 25, 21.5, 20.78, 25.9. $ R^2 $ test was also applied from \textit{sklearn.metrics} for checking the quality of fit for the linear regression. The average $ R^2 $ score of 0.39 was observed for simple linear regression, and for linear regression with PCA $ R^2 $ was 0.393. These results suggest that PCA has little to no effect on the prediction accuracy of linear regressor.

\begin{figure}[t]
	\centering
	\includegraphics[width = 8.5cm]{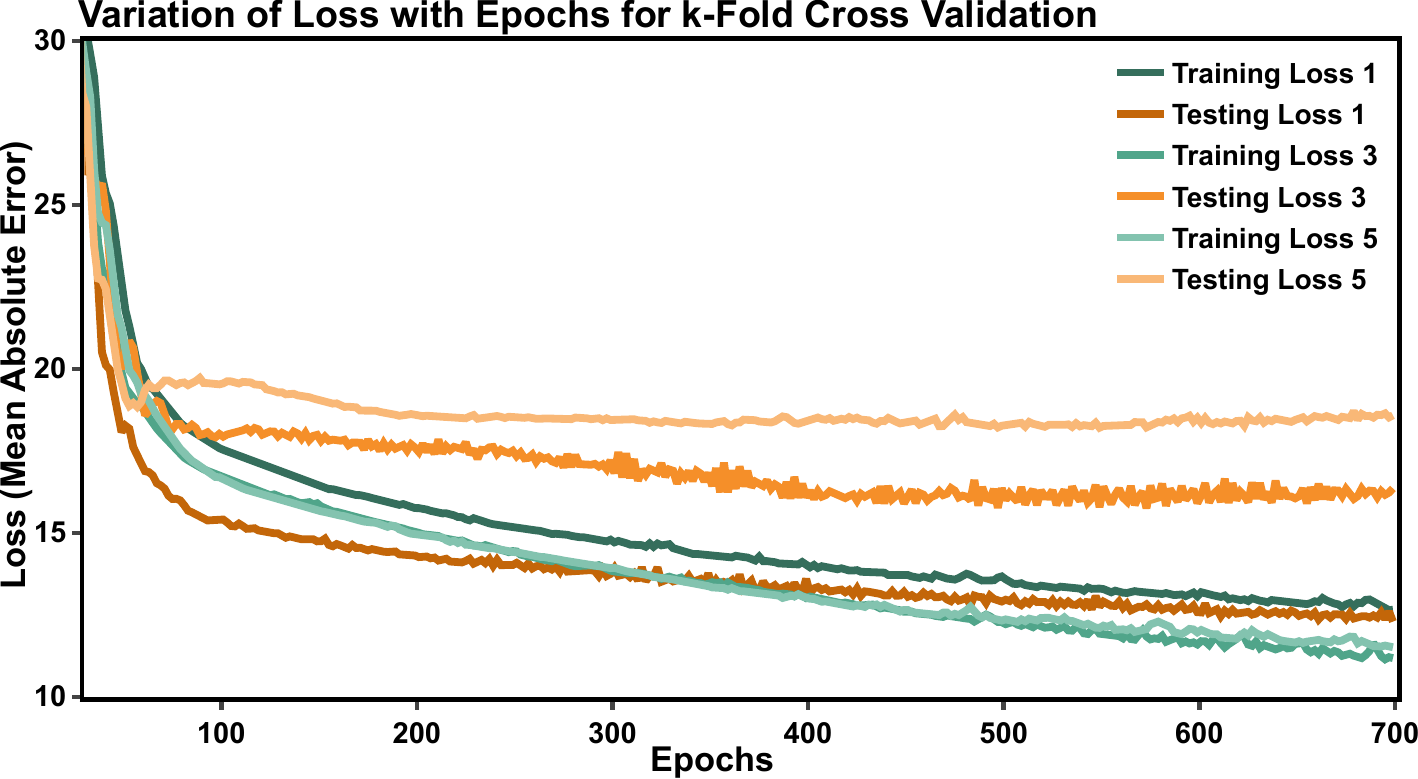}
	\caption{Variation of loss with epochs for different 5-fold sets used as training and validation sets.}
	\label{fig:maevsepoch}
\end{figure}

\subsubsection{Random Forest Regressor}
Random Forest is an ensemble learning method that combines multiple decision trees to make predictions \cite{breimanRandomForests2001}. It is a powerful and widely used algorithm in machine learning for both classification and regression tasks. Random Forest gets its name from the randomness introduced in two aspects: the random selection of features and the random sampling of training data \cite{hastieElementsStatisticalLearning2009}. \\ \textit{RandomForestRegressor} was used from \textit{sklearn.ensemble} library. The hyperparameters chosen for the training and testing have been shown in the following code snippet,
\begin{lstlisting}{python}
regressor = RandomForestRegressor(n_estimators = 20,
	criterion = "absolute_error",
\end{lstlisting}
In our case, the training MAE produced by the random forest model was 7.31, 6.45, 7.31, 7.18, 6.6 and testing accuracy was 12.71, 20.54, 17.33, 14.23, 17.87 without the application of PCA (see Figure \ref{lrcomp}c). The training MAE after the application of PCA was 6.9, 6.2, 6.32, 7.03, 6.7 and testing MAE was 13.53, 20.1, 16.92, 13.35, 17.97 for set 1-5 respectively (see Figure \ref{lrcomp}d).

\subsubsection{K-Nearest Neighbour Regressor}
k-Nearest Neighbors (k-NN) is another popular supervised machine learning algorithm used for both classification and regression tasks. It is a simple yet effective algorithm that makes predictions based on the similarity of data points in the feature space. For identifying the similarity between the data points, this algorithm calculates the Euclidean distance (or any other measure of distance) of one data point with k other data points. \textit{KNeighborsRegressor} class was used from \textit{sklearn.neighbor} library. Hyperparameters of the model were varied for better performance results and finally, we settled for the following,
\begin{lstlisting}{language=Python}
regressor = KNeighborsRegressor(n_neighbors=5,
			weights="uniform",
			algorithm="auto",
			leaf_size=30,
			metric="euclidean")
\end{lstlisting}
Without application of PCA, the training MAE with k-NN was 15.65, 13.47, 13.97, 15, 14.35 and testing MAE was 13.3, 19.21, 18.71, 15.67, 16.55 for set 1-5 respectively (see Figure \ref{lrcomp}e). On the other hand, after the application of PCA the training MAE was 16.85, 15.22, 15.3, 17.44, 15.74 and testing MAE was 18.21, 24.23, 20.5, 14.7, 19.35 respectively (see Figure \ref{lrcomp}f).

\subsubsection{Artificial Neural Network Regressor}
\begin{figure}
	\centering
	\includegraphics[width=8.5cm]{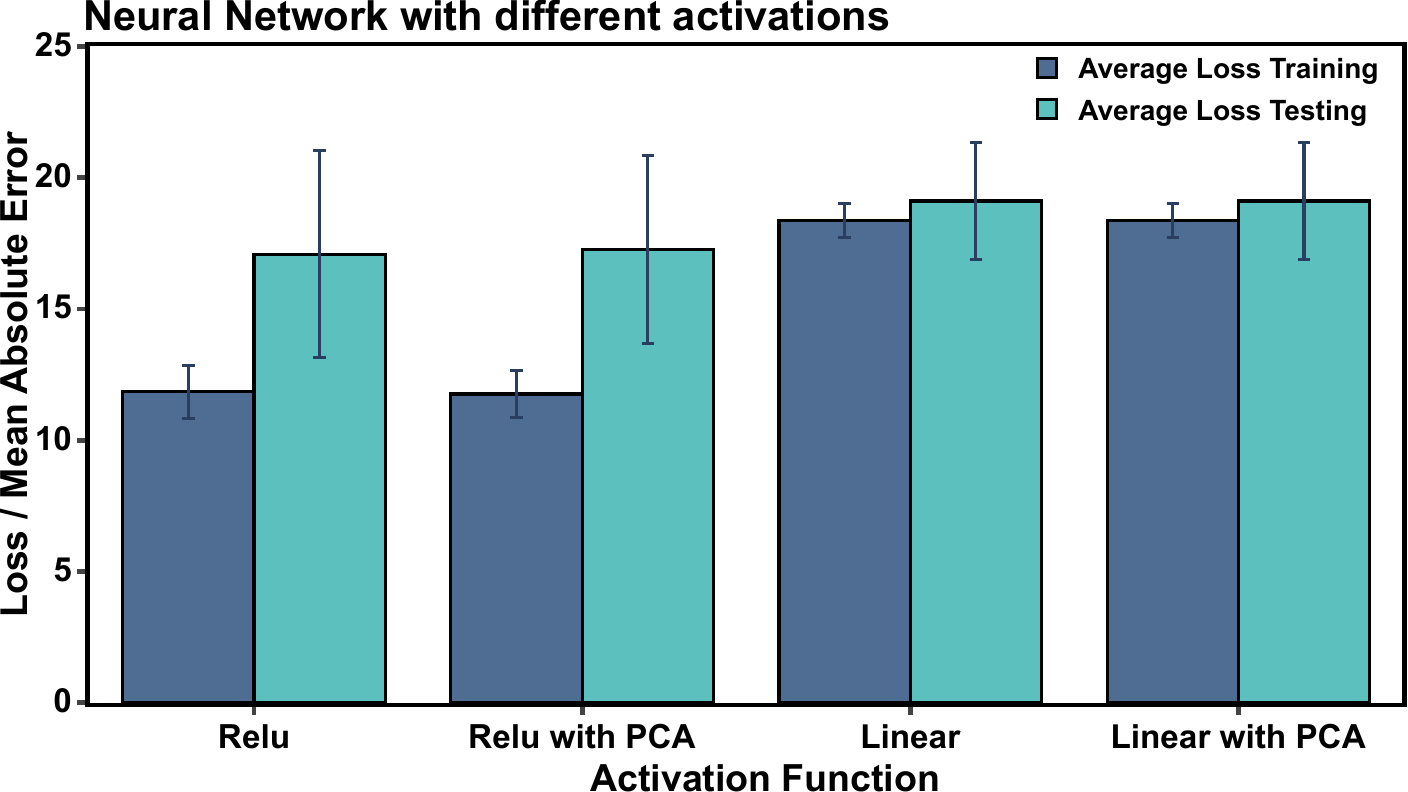}
	\caption{Comparison of mean absolute error between different activation functions with and without the application of principal component analysis.}
	\label{fig:pcaloss}
\end{figure}
A sequential artificial neural network was constructed using fully connected \textit{Dense} layers within a PyTorch back-end framework. The model architecture employed for prediction comprised a 3-500-500-3 configuration. The rectified linear unit (ReLU) activation function was selected due to its non-linear behavior \cite{agarapDeepLearningUsing2018}. An alternative activation function, "linear," was also evaluated but resulted in slightly inferior prediction outcomes (refer to Supplementary Figure \ref{supp-fig:kfoldact} c and \ref{fig:pcaloss}). The model underwent training for 700 epochs utilizing the adaptive momentum (Adam) optimizer and mean absolute error (MAE) as the loss function across all training and testing sets generated through 5-fold cross-validation. The Adam optimizer was configured with a learning rate of 0.001, an epsilon of $10^{-8}$, and $\beta$ values of (0.9, 0.999). The progression of MAE over the epochs is depicted in Supplementary Figure \ref{supp-fig:maevsepoch}. While the training loss exhibited a monotonic decline with the increasing number of epochs (see Supplementary Figure \ref{supp-fig:maevsepoch}), overtraining the model may lead to suboptimal generalization on validation data. The training MAE with ReLU activation across sets 1-5 without PCA were 12.66, 10.54, 11.19, 13.29, and 11.52, respectively. Correspondingly, the testing/validation MAE were 12.3, 23.75, 16.36, 14.45, and 18.59 (see Supplementary Figure \ref{supp-fig:kfoldact}a). Post-PCA application on both training and testing data, the training and testing MAE with ReLU activation were 12.88, 10.65, 11.32, 12.79, 11.15 and 13.3, 23.63, 16.03, 15.01, 18.31 across sets 1-5, respectively. The MAE exhibited a marginal increase upon PCA application to the training data, with insignificant effects on the testing data (see Supplementary Figure \ref{supp-fig:kfoldact}a and b). Conversely, the MAE with linear activation was generally higher than with ReLU, irrespective of PCA application (see Supplementary Figure \ref{supp-fig:kfoldact}c and d). The mean MAE for different ReLU/linear activations is illustrated in Figure \ref{fig:pcaloss} and Supplementary Figure \ref{supp-fig:kfoldact}.

\section{Discussion}

\subsection{Comparison of the Results}
\begin{figure}[h]
\centering
\includegraphics[width=8cm]{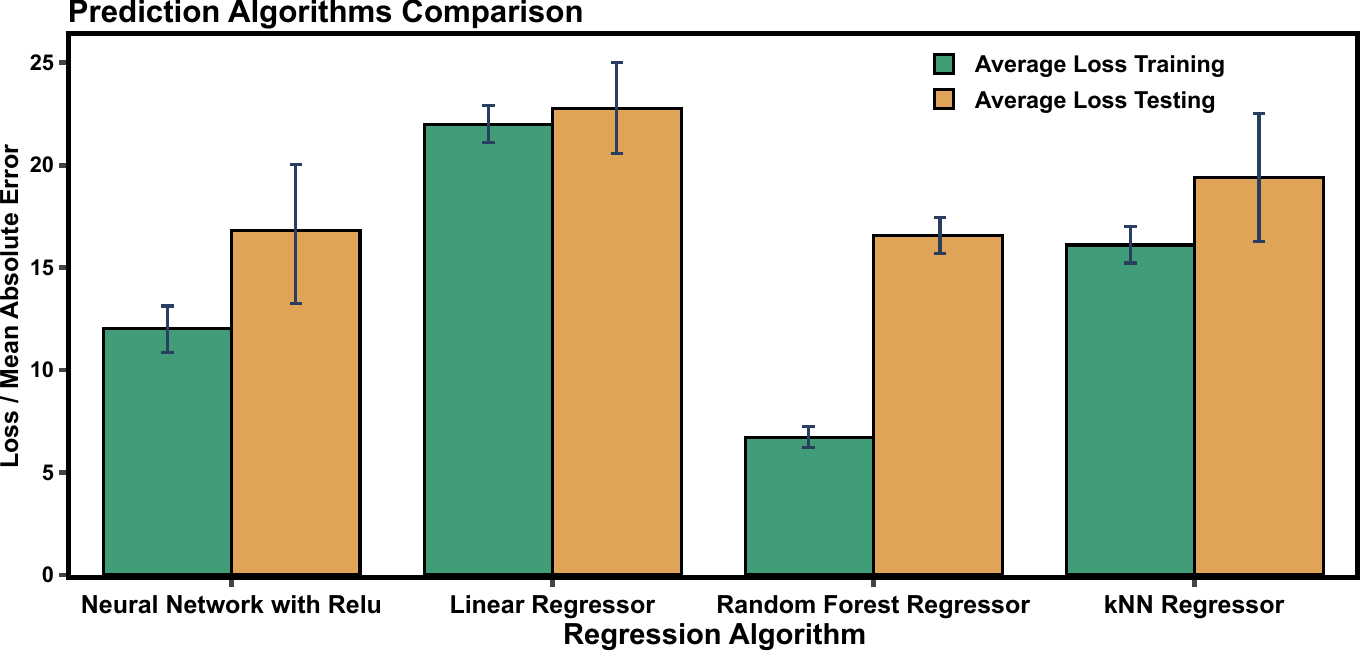}
\caption{This figure shows the comparison between the different algorithms used for the prediction of acid base concentration.}
\label{fig:comparison}
\end{figure}

Results of the best performances of the models have been taken from with or without PCA cases. These results were then compared with each other for fining out the two best algorithm for the prediction of acid base concentration in the solutions (see Figure \ref{fig:comparison}). 

The linear regressor produced Mean Absolute Error (MAE) values of $21.98 \pm 0.92$ and $22.8 \pm 2.2$ on the training and testing data, respectively. After applying PCA, these values remained unchanged.

For the kNN regressor, the mean MAE values were $14.5 \pm 0.77$ and $16.69 \pm 2.14$ on the training and testing data, respectively. After applying PCA, these values changed to $16.11 \pm 0.88$ and $19.4 \pm 3.1$.

For the Random Forest regressor, the mean MAE values were $6.97 \pm 0.37$ and $16.54 \pm 2.77$ on the training and testing data, respectively. After applying PCA, these values changed to $6.63 \pm 0.32$ and $16.37 \pm 2.6$.

Finally, the neural network showed the best performance with PCA and relu activation. The mean MAE values for training and testing were $11.76 \pm 0.9$ and $17.26 \pm 3.58$, respectively. 

The findings indicate that neural networks and random forest regressors exhibit superior performance in predicting acid-base concentrations in solutions. Consequently, these methodologies are recommended for forecasting the concentration of soil macronutrients based on soil physical parameters.
\subsection{Prediction of Soil Macronutrient Concentration}
For the prediction of soil nutrient concentrations, the soil samples were initially prepared for measurement (refer to the methods section). Subsequently, testing methodologies were employed to determine the pH, electrical conductivity, and average power metrics. A dataset was then compiled, encompassing the sample identifiers and the computed physical parameters. The two most effective machine learning models were utilized on this dataset to predict macronutrient concentrations. However, preliminary analyses indicated that the predicted concentrations deviated significantly from the empirically observed macronutrient levels. To rectify this inconsistency, a scaling factor was incorporated. The samples were bifurcated into two cohorts: the initial five samples were utilized to derive the mean scaling factor (see Table \ref{tab:scaling}), whereas the subsequent five samples were employed to evaluate the prediction error. This methodological adjustment was intended to enhance the accuracy of the model predictions, thereby aligning them more closely with the actual macronutrient concentrations.
\begin{table}
\caption{Calculated scaling factor for the used models.}
\label{tab:scaling}

\setlength{\tabcolsep}{9pt} 
\renewcommand{\arraystretch}{1.5} 
\centering
\begin{tabular}{|c|c|c|c|}

\hline \multicolumn{1}{|c|}{\textbf{}} & \multicolumn{1}{c|}{\textbf{\ce{HNO3}}} & \multicolumn{1}{c|}{\textbf{\ce{H3PO4}}} & \multicolumn{1}{c|}{\textbf{\ce{KOH}}} \\ \hline 
\cline{1-4}

\multicolumn{1}{|c|}{\textbf{Model}} & \multicolumn{3}{c|}{\textbf{Scaling Factor}}                                                \\ \cline{1-4}
\multicolumn{1}{|c|}{Neural Network} & \multicolumn{1}{c|}{NA} & \multicolumn{1}{c|}{0.054} & 0.234   \\ \cline{1-4}
\multicolumn{1}{|c|}{Random Forest}  & \multicolumn{1}{c|}{NA} & \multicolumn{1}{c|}{0.093} & 0.123   \\ \cline{1-4}

\end{tabular}
\end{table}

Upon comparing the results generated by the machine learning algorithms with the actual lab tests (see Table \ref{tab:error}), it was observed that the error in predicting \ce{P2O5} using the neural network was 26.3\%, while for \ce{K2O}, the error was 21.8\%. In contrast, the random forest regressor exhibited an error of 23.6\% for \ce{P2O5} and 16\% for \ce{K2O}. Although the random forest regressor appears to be more accurate based on these error rates, it failed to show any variation in the prediction of nutrient concentrations. Consequently, despite the higher error rates, the neural network regressor is considered a better algorithm for predicting the concentration of these nutrients due to its ability to capture variations in the data.

\begin{table}
\caption{Comparison of the results as predicted by the algorithms and the actual soil testing results.}
\label{tab:error}
\setlength{\tabcolsep}{2pt}
\renewcommand{\arraystretch}{1.5}
\centering
\begin{tabular}{|c|c|c|c|c|c|c|}

\hline
\multicolumn{1}{|c|}{} & \multicolumn{2}{c|}{\textbf{Lab Test}}            & \multicolumn{2}{c|}{\textbf{Neural Network}}      & \multicolumn{2}{c|}{\textbf{Random Forest}}       \\ \hline
\textbf{Sample} & \multicolumn{1}{c|}{\textbf{\ce{P2O5}}} & \textbf{\ce{K2O}} & \multicolumn{1}{c|}{\textbf{\ce{P2O5}}} & \textbf{\ce{K2O}} & \multicolumn{1}{c|}{\textbf{\ce{P2O5}}} & \textbf{\ce{K2O}} \\ \hline
\textbf{6}      & \multicolumn{1}{c|}{22.66}         & 279          & \multicolumn{1}{c|}{27.3}          & 207.2        & \multicolumn{1}{c|}{26.5}          & 212.3        \\ \hline
\textbf{7}      & \multicolumn{1}{c|}{22.66}         & 251          & \multicolumn{1}{c|}{27.4}          & 200.7        & \multicolumn{1}{c|}{26.5}          & 212.3        \\ \hline
\textbf{8}      & \multicolumn{1}{c|}{13.39}         & 157          & \multicolumn{1}{c|}{26.1}          & 235.5        & \multicolumn{1}{c|}{26.5}          & 212.3        \\ \hline
\textbf{9}      & \multicolumn{1}{c|}{20.6}          & 214          & \multicolumn{1}{c|}{28.0}          & 200.4        & \multicolumn{1}{c|}{26.2}          & 212.3        \\ \hline
\textbf{10}     & \multicolumn{1}{c|}{21.63}         & 220          & \multicolumn{1}{c|}{27.7}          & 201.4        & \multicolumn{1}{c|}{26.5}          & 212.3        \\ \hline

\multicolumn{3}{|c|}{\textbf{Percentage Error}} & \multicolumn{1}{c|}{26.3\%} & \multicolumn{1}{c|}{21.8\%} & \multicolumn{1}{c|}{23.6\%} & \multicolumn{1}{c|}{16\%} \\ \hline
\end{tabular}
\end{table}

\section{Conclusion}
According to the Food and Agriculture Organization (FAO), plant stress results in a global economic loss of approximately 30\%, valued at \$70 billion \cite{herforth2020cost}. A substantial portion of this loss can be alleviated through the provision of adequate nutrients to plants \cite{kumari2022plant,hasanuzzamanPotassiumVitalRegulator2018,kayaSulfurenrichedLeonarditeHumic2020a}. However, current methods such as chemical testing are labor-intensive and fail to provide real-time data on soil nutrient concentrations. To address this gap, this paper introduces a machine learning-based soil macronutrient monitoring approach that predicts nutrient concentrations utilizing soil physical parameters including conductivity, pH, and average power.

To train the machine learning algorithms, a dataset comprising physical properties such as conductivity, average power, and pH was curated. This dataset was developed by preparing artificial acid-base solutions that simulate variations in soil nutrients, followed by measuring the corresponding physical properties. Four distinct machine learning models—linear regressor, k-nearest neighbour, random forest, and neural network—were trained on this dataset. Among these, the two models that demonstrated superior predictive accuracy were selected. The random forest model exhibited prediction errors of 23.6\% for \ce{P2O5} and 16\% for \ce{K2O}. Conversely, the neural network model produced errors of 26.3\% and 21.8\% for predicting \ce{P2O5} and \ce{K2O}, respectively.

The method proposed in this paper necessitates minimal human intervention and can predict soil nutrient levels in real-time, thereby offering a proficient means to monitor soil and plant health. To further enhance model accuracy, it is imperative to expand the number of features in the dataset. For example, incorporating temperature as a feature could augment the model’s predictive capability, as temperature fluctuations affect both pH and conductivity. Furthermore, all the currently measured physical parameters are also affected by other nutrients such as magnesium, calcium, sulphur etc. Inclusion of these nutrients in the creation of soil phantoms would also boost the accuracy of the machine learning models. Also, increasing the dataset's sample size would further bolster the model’s accuracy and robustness.



{\appendix[Code for calculating the conductivity and average power]
\begin{lstlisting}[language=Python, caption = {}, label={code:imp1}][!h]


# Calculation for average power and conductivity.
N, P, K, pH = file[:-4].split("-")
data = pd.read_csv(file, header = None)
n = len(data[1]) - 1 #Number of samples for averaging.
Pav = round(simpson(data[1], data[0]), 7)

sigma = 0 #Initializing the sigma i.e. conductivity
for i in range(1, len(data[1]) - 1):
	#Calculation for the conductivity.
    cond = (data[1][i] - data[1][i - 1]) * \
    l / (0.05 * A * (n-1))
    sigma = sigma + cond
sigma = round(sigma, 7)


\end{lstlisting}


\bibliography{bibliography}

\begin{thebibliography}{10}
\providecommand{\url}[1]{#1}
\csname url@samestyle\endcsname
\providecommand{\newblock}{\relax}
\providecommand{\bibinfo}[2]{#2}
\providecommand{\BIBentrySTDinterwordspacing}{\spaceskip=0pt\relax}
\providecommand{\BIBentryALTinterwordstretchfactor}{4}
\providecommand{\BIBentryALTinterwordspacing}{\spaceskip=\fontdimen2\font plus
\BIBentryALTinterwordstretchfactor\fontdimen3\font minus
  \fontdimen4\font\relax}
\providecommand{\BIBforeignlanguage}[2]{{%
\expandafter\ifx\csname l@#1\endcsname\relax
\typeout{** WARNING: IEEEtran.bst: No hyphenation pattern has been}%
\typeout{** loaded for the language `#1'. Using the pattern for}%
\typeout{** the default language instead.}%
\else
\language=\csname l@#1\endcsname
\fi
#2}}
\providecommand{\BIBdecl}{\relax}
\BIBdecl

\bibitem{kimSoilMacronutrientSensing2009}
H.-J. Kim, K.~A. Sudduth, and J.~W. Hummel, ``Soil macronutrient sensing for
  precision agriculture,'' \emph{Journal of Environmental Monitoring}, vol.~11,
  no.~10, pp. 1810--1824, 2009.

\bibitem{hawkesford2023functions}
M.~J. Hawkesford, I.~Cakmak, D.~Coskun, L.~J. De~Kok, H.~Lambers, J.~K.
  Schjoerring, and P.~J. White, ``Functions of macronutrients,'' in
  \emph{Marschner's Mineral Nutrition of Plants}.\hskip 1em plus 0.5em minus
  0.4em\relax Elsevier, 2023, pp. 201--281.

\bibitem{lewis1986plants}
O.~A. Lewis and O.~A. Lewis, \emph{Plants and Nitrogen}.\hskip 1em plus 0.5em
  minus 0.4em\relax Cambridge University Press, 1986, no. 166.

\bibitem{raghothama2005phosphorus}
K.~G. Raghothama, ``Phosphorus and plant nutrition: An overview,''
  \emph{Phosphorus: Agriculture and the environment}, vol.~46, pp. 353--378,
  2005.

\bibitem{amtmann2012potassium}
A.~Amtmann and F.~Rubio, ``Potassium in plants,'' \emph{Els}, 2012.

\bibitem{hasanuzzamanPotassiumVitalRegulator2018}
M.~Hasanuzzaman, M.~B. Bhuyan, K.~Nahar, M.~S. Hossain, J.~A. Mahmud, M.~S.
  Hossen, A.~A.~C. Masud, and M.~Fujita, ``Potassium: A vital regulator of
  plant responses and tolerance to abiotic stresses,'' \emph{Agronomy}, vol.~8,
  no.~3, p.~31, 2018.

\bibitem{white2003calcium}
P.~J. White and M.~R. Broadley, ``Calcium in plants,'' \emph{Annals of botany},
  vol.~92, no.~4, pp. 487--511, 2003.

\bibitem{wangCriticalRolePotassium2013}
M.~Wang, Q.~Zheng, Q.~Shen, and S.~Guo, ``The critical role of potassium in
  plant stress response,'' \emph{International journal of molecular sciences},
  vol.~14, no.~4, pp. 7370--7390, 2013.

\bibitem{mu2021physiological}
X.~Mu and Y.~Chen, ``The physiological response of photosynthesis to nitrogen
  deficiency,'' \emph{Plant Physiology and Biochemistry}, vol. 158, pp. 76--82,
  2021.

\bibitem{aziz2014phosphorus}
T.~Aziz, M.~Sabir, M.~Farooq, M.~A. Maqsood, H.~R. Ahmad, and E.~A. Warraich,
  ``Phosphorus deficiency in plants: Responses, adaptive mechanisms, and
  signaling,'' \emph{Plant signaling: Understanding the molecular crosstalk},
  pp. 133--148, 2014.

\bibitem{machinandiarenaPotassiumPhosphitePrimes2012}
M.~F. Machinandiarena, M.~C. Lobato, M.~L. Feldman, G.~R. Daleo, and A.~B.
  Andreu, ``Potassium phosphite primes defense responses in potato against
  {{Phytophthora}} infestans,'' \emph{Journal of plant physiology}, vol. 169,
  no.~14, pp. 1417--1424, 2012.

\bibitem{oerkeCropLossesPests2006}
E.-C. Oerke, ``Crop losses to pests,'' \emph{The Journal of Agricultural
  Science}, vol. 144, no.~1, pp. 31--43, 2006.

\bibitem{kumarDecodingPhysiologicalResponse2023}
M.~Kumar, Z.~Saifi, and S.~D. Krishnananda, ``Decoding the physiological
  response of plants to stress using deep learning for forecasting crop loss
  due to abiotic, biotic, and climatic variables,'' \emph{Scientific Reports},
  vol.~13, no.~1, p. 8598, 2023.

\bibitem{bureauTotalMidyearPopulation2008}
U.~C. Bureau, \emph{Total Midyear Population for the World: 1950--2050.}\hskip
  1em plus 0.5em minus 0.4em\relax US Census Bureau Washington, DC, 2008.

\bibitem{jones2001laboratory}
J.~B. Jones, \emph{Laboratory Guide for Conducting Soil Tests and Plant
  Analysis}.\hskip 1em plus 0.5em minus 0.4em\relax CRC press, 2001.

\bibitem{khairaInductivelyCoupledPlasma2011}
{\relax BS}.~Khaira and G.~Gill, ``Inductively coupled plasma optical emission
  spectrometry ({{ICP-OES}}) in soil analysis,'' \emph{Journal of Analytical
  Chemistry}, vol.~66, no.~3, pp. 231--235, 2011.

\bibitem{nemesXrayFluorescenceSpectrometry2015}
A.~Nemes, B.~Simon, S.~Cs{\'a}nyi, and T.~Vigh, ``X-ray fluorescence
  spectrometry in soil analysis,'' \emph{Journal of Analytical Chemistry},
  vol.~70, no.~7, pp. 738--747, 2015.

\bibitem{kim2007simultaneous}
H.-J. Kim, J.~W. Hummel, K.~A. Sudduth, and P.~P. Motavalli, ``Simultaneous
  analysis of soil macronutrients using ion-selective electrodes,'' \emph{Soil
  Science Society of America Journal}, vol.~71, no.~6, pp. 1867--1877, 2007.

\bibitem{schlesingerBiogeochemistryAnalysisGlobal2013}
W.~H. Schlesinger and E.~S. Bernhardt, \emph{Biogeochemistry: An Analysis of
  Global Change}.\hskip 1em plus 0.5em minus 0.4em\relax Academic press, 2013.

\bibitem{marschnerMineralNutritionHigher1995}
H.~Marschner, ``Mineral nutrition of higher plants. 2nd (eds) {{Academic
  Press}},'' \emph{New York}, pp. 15--22, 1995.

\bibitem{olsen1954estimation}
S.~R. Olsen, \emph{Estimation of Available Phosphorus in Soils by Extraction
  with Sodium Bicarbonate}.\hskip 1em plus 0.5em minus 0.4em\relax US
  Department of Agriculture, 1954, no. 939.

\bibitem{garcia-pedrajasEmpiricalStudyBinary2011}
N.~{Garc{\'i}a-Pedrajas} and D.~{Ortiz-Boyer}, ``An empirical study of binary
  classifier fusion methods for multiclass classification,'' \emph{Information
  Fusion}, vol.~12, no.~2, pp. 111--130, 2011.

\bibitem{kumarNeuralNetworksClassroom2004}
S.~Kumar, \emph{Neural {{Networks}}: {{A Classroom Approach}}}, ser. Computer
  Engineering Series.\hskip 1em plus 0.5em minus 0.4em\relax McGraw-Hill
  Education (India) Pvt Limited, 2004.

\bibitem{eash2015soil}
N.~S. Eash, T.~J. Sauer, D.~O'Dell, and E.~Odoi, \emph{Soil Science
  Simplified}.\hskip 1em plus 0.5em minus 0.4em\relax John Wiley \& Sons, 2015.

\bibitem{jones2012plant}
J.~B. Jones~Jr, \emph{Plant Nutrition and Soil Fertility Manual}.\hskip 1em
  plus 0.5em minus 0.4em\relax CRC press, 2012.

\bibitem{mustafa2011characterization}
{\relax AA}.~Mustafa, M.~Singh, {\relax NAYAN}.~Ahmed, {\relax RN}.~Sahoo,
  {\relax MANOJ}.~Khanna, A.~Sarangi, and {\relax AK}.~Mishra,
  ``Characterization and classification of soils of kheragarah, agra and their
  productivity potential,'' \emph{Journal of Water Management}, vol.~19, no.~1,
  pp. 1--19, 2011.

\bibitem{pauline1946flame}
{\relax BY}.~Pauline and M.~Hald, ``The flame photometer for the measurement of
  sodium and potassium in biological materials,'' \emph{The Journal of
  biological chemistry}, vol. 167, no.~2, pp. 499--510, 1946.

\bibitem{montgomery2021introduction}
D.~C. Montgomery, E.~A. Peck, and G.~G. Vining, \emph{Introduction to Linear
  Regression Analysis}.\hskip 1em plus 0.5em minus 0.4em\relax John Wiley \&
  Sons, 2021.

\bibitem{breimanRandomForests2001}
L.~Breiman, ``Random forests,'' \emph{Machine learning}, vol.~45, pp. 5--32,
  2001.

\bibitem{hastieElementsStatisticalLearning2009}
T.~Hastie, R.~Tibshirani, J.~H. Friedman, and J.~H. Friedman, \emph{The
  Elements of Statistical Learning: Data Mining, Inference, and
  Prediction}.\hskip 1em plus 0.5em minus 0.4em\relax Springer, 2009, vol.~2.

\bibitem{agarapDeepLearningUsing2018}
A.~F. Agarap, ``Deep learning using rectified linear units (relu),''
  \emph{arXiv preprint arXiv:1803.08375}, 2018.

\bibitem{herforth2020cost}
A.~Herforth, Y.~Bai, A.~Venkat, K.~Mahrt, A.~Ebel, and W.~A. Masters,
  \emph{Cost and Affordability of Healthy Diets across and within Countries:
  {{Background}} Paper for {{The State}} of {{Food Security}} and {{Nutrition}}
  in the {{World}} 2020. {{FAO Agricultural Development Economics Technical
  Study No}}. 9}.\hskip 1em plus 0.5em minus 0.4em\relax Food \& Agriculture
  Org., 2020, vol.~9.

\bibitem{kumari2022plant}
V.~V. Kumari, P.~Banerjee, V.~C. Verma, S.~Sukumaran, M.~A.~S. Chandran, K.~A.
  Gopinath, G.~Venkatesh, S.~K. Yadav, V.~K. Singh, and N.~K. Awasthi, ``Plant
  nutrition: {{An}} effective way to alleviate abiotic stress in agricultural
  crops,'' \emph{International Journal of Molecular Sciences}, vol.~23, no.~15,
  p. 8519, 2022.

\bibitem{kayaSulfurenrichedLeonarditeHumic2020a}
C.~Kaya, M.~{\c S}enbayram, N.~A. Akram, M.~Ashraf, M.~N. Alyemeni, and
  P.~Ahmad, ``Sulfur-enriched leonardite and humic acid soil amendments enhance
  tolerance to drought and phosphorus deficiency stress in maize ({{Zea}} mays
  {{L}}.),'' \emph{Scientific Reports}, vol.~10, no.~1, p. 6432, Dec. 2020.

\end{thebibliography}
\bibliographystyle{IEEEtran}




\vspace{-30pt}


\vspace{-30pt}


\vspace{-30pt}


\vfill

\end{document}


\maketitle
\thispagestyle{empty}
\begin{figure*}[!h]
	\centering
	\includegraphics[width=14cm]{Figures/Neural Networks K Fold}
	\caption{This figure contains the K-Fold cross validation with different activations and the application of PCA. a. K-Fold cross validation with ReLU activation. b. ReLU with PCA. c. K-Fold cross validation with linear activation. d. Linear activation with PCA.}
	\label{fig:kfoldact}
\end{figure*}

\begin{figure*}[!ht]
	\centering
	\includegraphics[width=14cm]{Figures/Epochs vs MAE}
	\caption{This figure shows the variation of MAE with epochs for all the cases of artificial neural networks.}
	\label{fig:maevsepoch}
\end{figure*}